\documentclass[letterpaper]{article}
\usepackage{aaai}
\usepackage{times}
\usepackage{helvet}
\usepackage{graphicx}
\usepackage{courier}
\usepackage{verbatim}
\usepackage{multirow}
\usepackage{wrapfig}
\usepackage{caption}
\usepackage{subcaption}
\begin{document}
%
\title{Automating Direct Speech Variations in Stories and Games}

\author{Stephanie M. Lukin \and James O. Ryan \and Marilyn A. Walker\\
Natural Language and Dialogue Systems Lab\\
University of California, Santa Cruz \\
1156 High Street, Santa Cruz, CA 95060 \\
{\tt \{slukin,joryan,mawalker\}@ucsc.edu}}
\maketitle
\begin{abstract}
\begin{quote}

Dialogue authoring in large games requires not only content creation
but the subtlety of its delivery, which can vary from character to
character. Manually authoring this dialogue can be tedious,
time-consuming, or even altogether infeasible. This paper 
utilizes a rich narrative representation for modeling dialogue and an
expressive natural language generation engine for realizing it, and 
expands upon a translation tool that bridges the two.
We add
functionality to the translator to allow direct speech to be modeled
by the narrative representation, whereas the original translator
supports only narratives told by a third person narrator. We show that
we can perform character substitution in dialogues. We implement and
evaluate a potential application to dialogue implementation:
generating dialogue for games with big, dynamic, or
procedurally-generated open worlds. We present a pilot study on human
perceptions of the 
personalities of characters using direct speech, assuming unknown
personality types at the time of authoring.
\end{quote}
\end{abstract}

\noindent 
Dialogue authoring in large games requires not only the creation of
content, but the subtlety of its delivery which can vary from
character to character. Manually authoring this dialogue can be
tedious, time-consuming, or even altogether infeasible. The task
becomes particularly intractable for games and stories with dynamic
open worlds in which character parameters that should produce
linguistic variation may change during gameplay or are decided
procedurally at runtime. Short of writing all possible variants
pertaining to all possible character parameters for all of a game's
dialogue segments, authors working with highly dynamic systems
currently have no recourse for producing the extent of content that
would be required to account for all linguistically meaningful
character states. As such, we find open-world games today filled with
stock dialogue segments that are used repetitively by many characters
without any linguistic variation, even in game architectures
with rich character models
that could give an actionable account of how
their speech may vary \cite{klabunde2013greetings}. 

Indeed, in general, we are building computational systems that,
underlyingly, are far more expressive than can be manifested by
current authoring practice. These concerns can also be seen in linear games,
in which the number of story paths may be limited to reduce
authoring time or which may require a large number of authors to create a
variety of story paths. Recent work explores the introduction of
automatically authored dialogues using expressive natural language
generation (NLG) engines, thus allowing for more content
creation and the potential of larger story paths
\cite{montfort2014expressing,lin2011all,cavazza2005dialogue,rowe2008archetype}.

\begin{figure}[thb!]
\begin{center}
\includegraphics[width=3.0in]{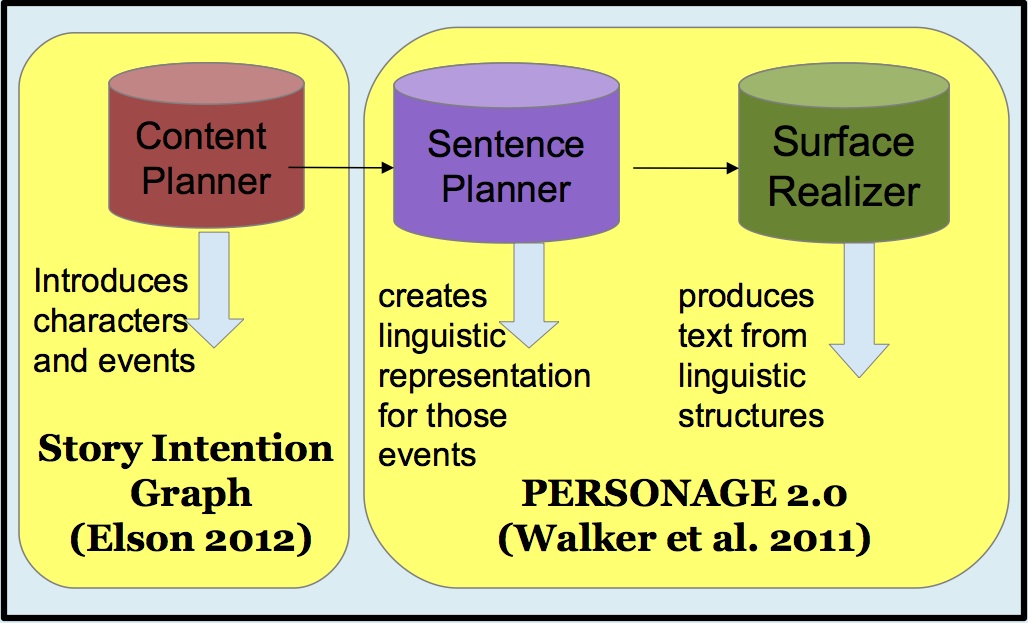}
\caption{\label{est-arch-fig} {NLG pipeline method of the ES Translator.}}
\end{center}
\end{figure}

\cite{walker2013using} explore using a dynamic and customizable NLG
engine called {\sc personage} 
to generate a variety of character styles and realizations,
as one way to help authors to reduce the
authorial burden of writing dialogue instead of relying on
scriptwriters.
{\sc personage} is a parameterizable NLG
engine grounded in the Big Five personality traits that provides a larger
range of pragmatic and stylistic variations of a single utterance than
other NLG engines \cite{MairesseWalker11}. In {\sc personage},
narrator's voice (or style to be conveyed) is controlled by a model 
that specifies values for different stylistic parameters (such as
verbosity, syntactic complexity, and lexical choice). {\sc personage}
requires hand crafted text plans, limiting not only the expressibility
of the generations, but also the domain.

\begin{figure*}[th!]
\centering                               
\begin{small}
\begin{tabular}{|p{3.3in}|p{3.3in}|}
\hline
Original Fable & Dialogic Interpretation of Original Fable \\ \hline
A Crow was sitting on a branch of a tree with a piece of cheese in her beak when a Fox observed her and set his wits to work to discover some way of getting the cheese.
Coming and standing under the tree he looked up and said, ``What a noble bird I see above me! Her beauty is without equal, the hue of her plumage exquisite. If only her voice is as sweet as her looks are fair, she ought without doubt to be Queen of the Birds.''
The Crow was hugely flattered by this, and just to show the Fox that she could sing she gave a loud caw.
Down came the cheese,of course, and the Fox, snatching it up, said, ``You have a voice, madam, I see: what you want is wits.''
&
``It's a lovely day, I think I will eat my cheese here'' the crow said, flying to a branch with a piece of cheese in her beak. 
A Fox observed her. ``I'm going to set my wits to work to discover some way to get the cheese''
Coming and standing under the tree he looked up and said, ``What a noble bird I see above me! Her beauty is without equal, the hue of her plumage exquisite. If only her voice is as sweet as her looks are fair, she ought without doubt to be Queen of the Birds.''
``I am hugely flattered!'' said the Crow. ``Let me sing for you!''
Down came the cheese,of course, and the Fox, snatching it up, said, ``You have a voice, madam, I see: what you want is wits." \\ \hline
\end{tabular}
\caption{\label{fc-original}The Fox and The Crow}
\end{small}
\end{figure*}

\cite{reed2011step} introduce SpyFeet: a mobile game to encourage
physical activity which makes use of dynamic storytelling and
interaction. A descendant of {\sc personage}, called SpyGen, is its
NLG engine. The input to SpyGen is a text plan from Inform7, which
acts as the content planner and manager.
\cite{reed2011step} show that this architecture allows any character
personality to be used in any game situation. However their approach
was not evaluated and it relied on game specific text plans.

\cite{rishes2013generating} created a translator, called the
ES-Translator ({\sc est}), which bridges a narrative representation
produced by the annotation tool Scheherezade, to the representation
required by {\sc personage}, thus not requiring the creation of text
plans. Fig.~\ref{est-arch-fig}
provides a high level view of the architecture of {\sc est}, described
in more detail below. Scheherazade annotation facilitates the creation of a rich
symbolic representation for narrative texts, using a schema known as
the {\sc story intention graph} or {\sc sig}
\cite{ElsonMcKeown10,elson2012b}. A {\sc sig} represents the sequence
of story events, as well as providing a rich representation of the
intentions, beliefs, and motivations of story characters. The {\sc est} takes the {\sc sig} as input, and then converts the narrative
into a format that {\sc personage} can utilize.

However, the approach described in \cite{rishes2013generating} is
limited to telling stories from the third person narrator
perspective. This paper expands upon the {\sc est} to enable
annotation of direct speech in Scheherazade, that can then be realized
directly as character dialogue. We explore and implement a potential
application to producing dialogue in game experiences for large,
dynamic, or procedurally-generated open worlds, and present a pilot study
on user perceptions of the personalities of story characters 
who use direct speech. 
The contributions of this work are: 1) we have can modify a single, underlying 
representation of narrative to adjust for direct speech and substitute character 
speaking styles; and 2) that we can perform this modeling on any domain.

\section{ES Translator}

Aesop's Fable ``The Fox and The Crow" (first column in
Fig.~\ref{fc-original}) is used to illustrate the development
and the new dialogue expansion of the {\sc est}.

\subsection{Annotation Schema}

\begin{figure*}[tb]
\begin{center}
\includegraphics[width=4.0in]{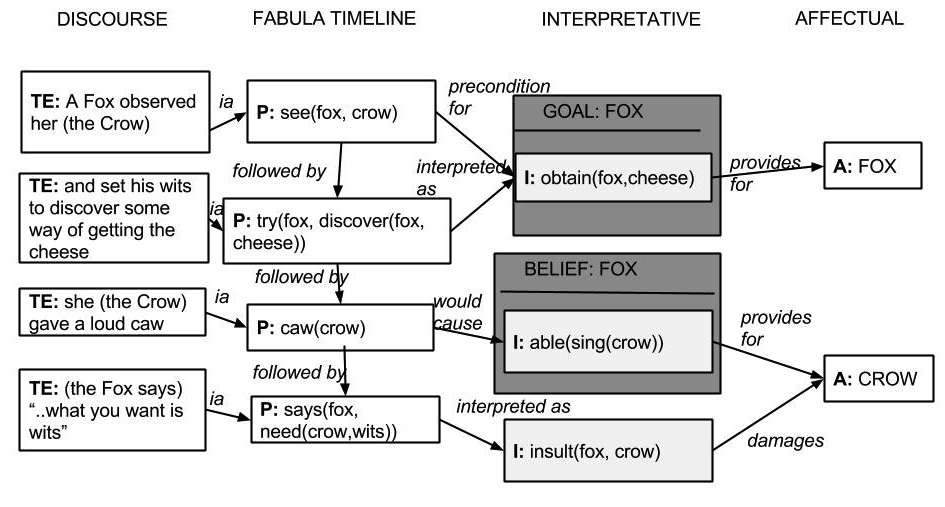}
\vspace{-0.2in}
\caption{\label{foxcrow-sig} Part of the {\sc Story Intention Graph} ({\sc sig}) for ``The Fox and The Crow"}
\end{center}
\end{figure*}

\begin{figure}[htb]
\begin{center}
\includegraphics[width=0.4\textwidth]{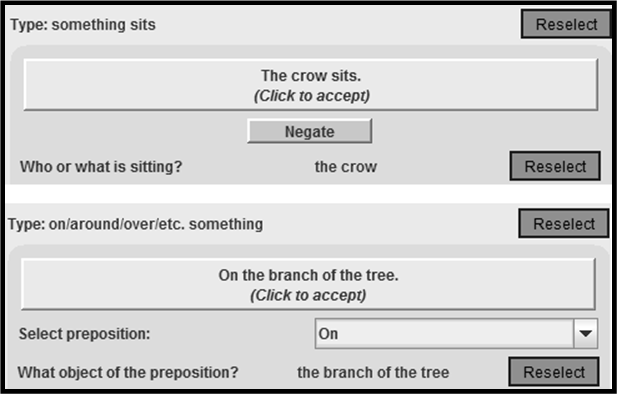}
\caption{\label{gui} GUI view of propositional modeling}
\vspace{-0.2in}
\end{center}
\end{figure}

One of the strengths of Scheherazade is that it allows users to annotate a story along several dimensions, starting with the surface form of the story (first column in Fig.~\ref{foxcrow-sig}) and then proceeding to deeper representations.  The first dimension (second column in Fig.~\ref{foxcrow-sig}) is called the ``timeline  layer'', in which the story facts are encoded as predicate-argument structures (propositions) and temporally ordered on a timeline. The timeline layer consists of a network of propositional structures, where nodes correspond to lexical items that are linked by thematic relations. Scheherazade adapts information about predicate-argument structures from the VerbNet lexical database \cite{Kipper06} and uses WordNet \cite{WordNet} as its noun and adjectives taxonomy. The arcs of the story graph are labeled with discourse relations. Fig.~\ref{gui} shows a GUI screenshot of assigning propositional structure to the sentence {\it The crow was sitting on the branch of a tree}. This sentence is encoded as two nested propositions {\tt sit(crow)} and the prepositional phrase {\tt on(the branch of the tree)}. Both actions ({\tt sit} and {\tt on}) contain references to the story characters and objects ({\tt crow} and {\tt branch of the tree}) that fill in slots corresponding to semantic roles. Only the timeline layer is utilized for this work at this time. 

In the current annotation tool, the phrase {\it The fox ... said ``You have a voice, madam..."} can be annotated in Scheherazade by selecting {\tt say} from VerbNet and attaching the proposition {\it the crow has a voice} to the verb {\tt say(fox, able-to(sing(crow)))}. However, this is realized as {\it The fox said the crow was able to sang} (note: in the single narrator realization, everything is realized in the past tense at this time. When we expand to direct speech in this work, we realize verbs in the future or present tense where appropriate). To generate {\it The fox said ``the crow is able to sing"}, we append the modifier ``directly" to the verb ``say" (or any other verb of communication or cognition, e.g. ``think"), then handle it appropriately in the {\sc est} rules described in Section~\ref{est-rules}. Furthermore, to generate {\it The fox said ``you are able to sing"}, instead of selecting {\tt crow}, an {\tt interlocutor} character is created and then annotated as {\tt say(fox, able-to(sing(interlocutor)))}. We add new rules to the {\sc est} to handle this appropriately.


\subsection{Translation Rules}
\label{est-rules}

\begin{figure*}[th!]
\centering
\includegraphics[width=5.0in]{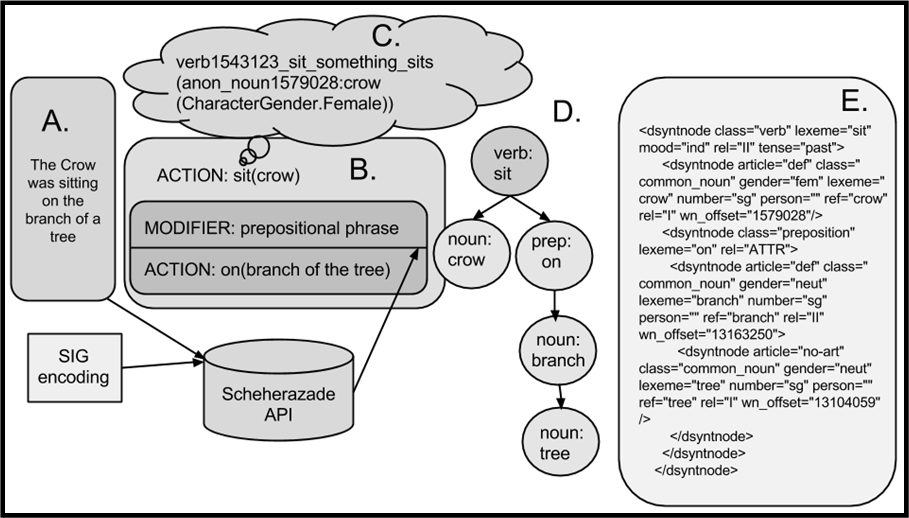}
\vspace{-0.1in}
\caption{\label{sig_dsynts} Step by step transformation from {\sc sig} to DSyntS}
\vspace{-0.2in}
\end{figure*}

The process of the {\sc est} tranformation of the {\sc sig} into a
format that can be used by {\sc personage} is a multi-stage process
shown in Fig.~\ref{sig_dsynts} \cite{rishes2013generating}.  First, a
syntactic tree is constructed from the propositional event
structure. Element A in Fig.~\ref{sig_dsynts} contains a sentence from
the original ``The Fox and the Grapes" fable.  The Scheherazade API is
used to process the fable text together with its {\sc sig} encoding
and extract actions associated with each timespan of the timeline
layer.  Element B in Fig.~\ref{sig_dsynts} shows a schematic
representation of the propositional structures.  Each action
instantiates a separate tree construction procedure.  For each action,
we create a verb instance (highlighted nodes of element D in
Fig.~\ref{sig_dsynts}). Information about the predicate-argument frame
that the action invokes (element C in Fig.~\ref{sig_dsynts}) is then
used to map frame constituents into respective lexico-syntactic
classes, for example, characters and objects are mapped into nouns,
properties into adjectives and so on. The lexico-syntactic class
aggregates all of the information that is necessary for generation of
a lexico-syntactic unit in the DSyntS representation used by the {\sc
  realpro} surface realizer of {\sc personage} (element E in
Fig.~\ref{sig_dsynts})
\cite{lavoie1997fast}. \cite{rishes2013generating} define 5 classes
corresponding to main parts of speech: noun, verb, adverb, adjective,
functional word. Each class has a list of properties such as
morphology or relation type that are required by the DSyntS notation
for a correct rendering of a category. For example, all classes
include a method that parses frame type in the {\sc sig} to derive the
base lexeme. The methods to derive grammatical features are
class-specific. Each lexico-syntactic unit refers to the elements that
it governs syntactically thus forming a hierarchical structure. A
separate method collects the frame adjuncts as they have a different
internal representation in the {\sc sig}.

At the second stage, the algorithm traverses the syntactic tree in-order
and creates an XML node for each lexico-syntactic unit. Class
properties are then written to disk, and the resulting file (see element E in Fig.~\ref{sig_dsynts}) is processed by the surface realizer to generate text.

\subsection{Dialogue Realization}

The main advantage of {\sc personage} is its ability to generate a
single utterance in many different voices.  Models of narrative style
are currently based on the Big Five personality traits
\cite{MairesseWalker11}, or are learned from film scripts
\cite{Walkeretal11}. Each type of model (personality trait or film)
specifies a set of language cues, one of 67 different parameters,
whose value varies with the personality or style to be
conveyed. In \cite{reed2011step}, the SpyGen engine was not evaluated.
However previous work \cite{MairesseWalker11} has shown that humans
perceive the personality stylistic models in the way that {\sc
  personage} intended, and \cite{Walkeretal11} shows that character
utterances in a new domain can be recognized by humans
as models based on a particular film character.

Here we first show that our new architecture as illustrated by Fig.~\ref{est-arch-fig} and
Fig.~\ref{sig_dsynts} lets us develop {\sc sig}s for any content domain.
We first illustrate how we can change domains
to a potential game dialogue where the player could have a choice of
party members, and show that the {\sc est} is capable of such
substitutions. Table~\ref{game} shows different characters saying the
same thing in their own style. We use an openness to experience model
from the Big 5 \cite{MairesseWalker11}, Marion from {\it Indiana Jones} and Vincent from {\it
  Pulp Fiction} from \cite{lin2011all}, and the Otter character model
from \cite{reed2011step}'s Heart of Shadows. 

\begin{table}[h]
\centering
\caption{\label{game}Substituting Characters}
\begin{small}
\centering
\begin{tabular}{|r|p{1.8in}|}
\hline
Openness (Big Five) & ``Let's see... I see, I will fight with you, wouldn't it? It seems to me that you save me from the dungeon, you know." \\ \hline
Marion (Indiana Jones) & ``Because you save me from the dungeon pal, I will fight with you!" \\ \hline
Vincent (Pulp Fiction) & ``Oh God I will fight with you!" \\ \hline
Otter (Heart of Shadows) & ``Oh gosh I will fight with you because you save me from the dungeon mate!" \\ 
\hline
\end{tabular}
\end{small}
\end{table}

With the {\sc est}, an author could use Scheherazade to encode stock
utterances that any character may say, and then have {\sc personage}
automatically generate stylistic variants of that utterance pertaining
to all possible character personalities. 
This technique would be particularly ripe for games in which character
personality is unknown at the time of authoring. In games like this,
which include \textit{The Sims 3} \cite{sims3} and \textit{Dwarf
  Fortress} \cite{dwarffortress}, personality may be dynamic or
procedurally decided at runtime, in which case a character could be
assigned any personality from the space of all possible personalities
that the game can model. Short of writing variants of each dialogue
segment for each of these possible personalities, authors for games
like these simply have no recourse for producing enough dialogic
content sufficient to cover all linguistically meaningful character
states. For this reason, in these games a character's personality does
not affect her dialogue. Indeed, \textit{The Sims 3} avoids
natural-language dialogue altogether, and \textit{Dwarf Fortress},
with likely the richest personality modeling ever seen in a game,
features stock dialogue segments that are used across all characters,
regardless of speaker personality.

When producing the {\sc est} \cite{rishes2013generating} focused on a
tool that could generate variations of Aesop's Fables such as {\it The
  Fox and the Crow} from Drama Bank~ \cite{ElsonMcKeown10}.  
\cite{rishes2013generating} evaluated the {\sc est} with a primary
focus on whether the {\sc est} produces {\bf correct} retellings of
the fable.  They measure the generation produced by the {\sc est} in
terms of the string similarity metrics BLEU score and Levenshtein
Distance to show that the new realizations are comparable to the
original fable.

After we add new rules to the {\sc est} for handling direct speech and
interlocutors, we modified the original {\sc sig} representation of
the {\it Fox and the Crow} to contain more dialogue in order to
evaluate a broader range of character styles, along with the use of direct speech (second column of
Fig.~\ref{fc-original}). This version is annotated using the new
direct speech rules, then run through the {\sc est} and {\sc
  personage}.  Table~\ref{pers-fig} shows a subset of parameters,
which were used in the three personality models we tested here: the
{\it laid-back} model for the fox's direct speech, the {\it shy} model
for the crow's direct speech, and the {\it neutral} model for the
narrator voice. The {\it laid-back} model uses emphasizers, hedges,
exclamations, and expletives, whereas the {\it shy} model uses
softener hedges, stuttering, and filled pauses. The {\it neutral}
model is the simplest model that does not utilize any of the extremes
of the {\sc personage} parameters.

\begin{table}[!htb]
\begin{scriptsize}
\begin{tabular}{|p{0.35in}p{0.55in}p{.95in}p{.75in}|c|c|c|c|}
\hline
{\bf Model} & {\bf Parameter} & {\bf Description} & {\bf Example}\\
\hline
\multirow{3}{*}{Shy} & {\sc Softener hedges} & Insert syntactic elements ({\it sort of},
{\it  kind of}, {\it somewhat}, {\it quite}, {\it around}, {\it
rather}, {\it I think that}, {\it it seems that}, {\it it seems to
me that}) to mitigate the strength of a proposition & {\it `It seems to me that he was hungry'}\\ 

&{\sc Stuttering} &  Duplicate parts of a content word & {\it `The vine hung on the tr-trellis'}\\

&{\sc Filled pauses} & Insert  syntactic elements expressing
hesitancy ({\it I  mean},  {\it err},  {\it mmhm}, {\it like}, {\it
you~know}) & {\it `Err... the fox jumped'}\\
\hline
\multirow{3}{*}{Laid-back} & {\sc Emphasizer hedges} & Insert  syntactic elements ({\it really},
{\it basically}, {\it actually}) to strengthen a
proposition & {\it `The fox failed to get the group of grapes, alright?'}  \\ 

& {\sc Exclamation} & Insert an exclamation mark & {\it `The group of grapes hung on the vine!'} \\

& {\sc Expletives} & Insert  a swear word & {\it `The fox was damn hungry'}\\
\hline
\end{tabular}
\end{scriptsize}
\vspace{-0.1in}
\centering \caption{\label{pers-fig} Examples of pragmatic marker
insertion parameters from {\sc personage}}
\vspace{-0.1in}
\end{table}

\begin{figure*}[htb]
\centering
\begin{small}
\begin{tabular}{|p{3.1in}|p{3.7in}|}
\hline
Single Narrator Realization & Dialogic Realization \\ \hline
The crow sat on the branch of the tree.
       The cheese was in the beak of the crow.
       The fox observed the crow.
       The fox tried he discovered he got the cheese.
       The fox came.
       The fox stood under the tree.
       The fox looked toward the crow.
       The fox said he saw the bird.
       The fox said the beauty of the bird was incomparable.
       The fox said the hue of the feather of the bird was exquisite.
       The fox said if the pleasantness of the voice of the bird was equal to the comeliness of the appearance of the bird the bird undoubtedly was every queen of every bird.
       The crow felt the fox flattered her.
       The crow loudly cawed in order for her to showed she was able to sang.
       The cheese fell.
       The fox snatched the cheese.
       The fox said the crow was able to sang.
       The fox said the crow needed the wits. 
&
The crow sat on the tree's branch.
The cheese was in the crow's pecker.
The crow thought ``I will eat the cheese on the branch of the tree because the clarity of the sky is so-somewhat beautiful."
The fox observed the crow.
The fox thought ``I will obtain the cheese from the crow's nib."
The fox came.
The fox stood under the tree.
The fox looked toward the crow.
The fox avered ``I see you!"
The fox alleged `your's beauty is quite incomparable, okay?"
The fox alleged `your's feather's chromaticity is damn exquisite."
The fox said ``if your's voice's pleasantness is equal to your's visual aspect's loveliness you undoubtedly are every every birds's queen!"
The crow thought ``the fox was so-somewhat flattering."
The crow thought ``I will demonstrate my's voice."
The crow loudly cawed.
The cheese fell.
The fox snatched the cheese.
The fox said ``you are somewhat able to sing, alright?"
The fox alleged ``you need the wits!"\\ \hline
\end{tabular}
\caption{\label{fc-est}The Fox and The Crow {\sc est} Realizations}
\end{small}
\end{figure*}

We first illustrate a monologic version of ``The Fox and The Crow" as produced by the {\sc est} in the first column of Table~\ref{fc-est}. This is our baseline realization. The second column shows the {\sc ets}'s realization of the fable encoded in dialogue with the models described above. 

We run {\sc personage} three times, one for each of our {\sc personage} models ({\it laid-back}, {\it shy}, and {\it neutral}), then have a script that selects the narrator realization by default, and in the event of a direct speech instance, piece together realizations from the crow or the fox. We are currently exploring modifications to our system that allows multiple personalities to be loaded and assigned to characters so that {\sc personage} is only run once and the construction be automated. Utterances are generated in real-time, allowing the underlying {\sc personage} model to change at any time, for example, to reflect the mood or tone of the current situation in a game.

\section{User Perceptions}
\label{apps}

Here we present a pilot study aimed at illustrating how the
flexibility of the {\sc est} when producing dialogic variations allows
us to manipulate the perception of the story characters.  We collect
user perceptions of the generated dialogues via an experiment on
Mechanical Turk in which the personality models used to generate the
dialogic version of ``The Fox and The Crow" shown in Fig.~\ref{fc-est}
are modified, so that the fox uses the {\it shy} model and the crow
uses the {\it laid-back} model. We have three conditions; participants
are presented with the dialogic story told 1) only with the {\it
  neutral} model; 2) with the crow with {\it shy} and the fox with
{\it laid-back}; and 3) with the crow with {\it laid-back} and the fox
with {\it shy}.

\begin{figure*}[tbh]
        \centering
        \begin{subfigure}[b]{0.3\textwidth}
                \includegraphics[width=\textwidth]{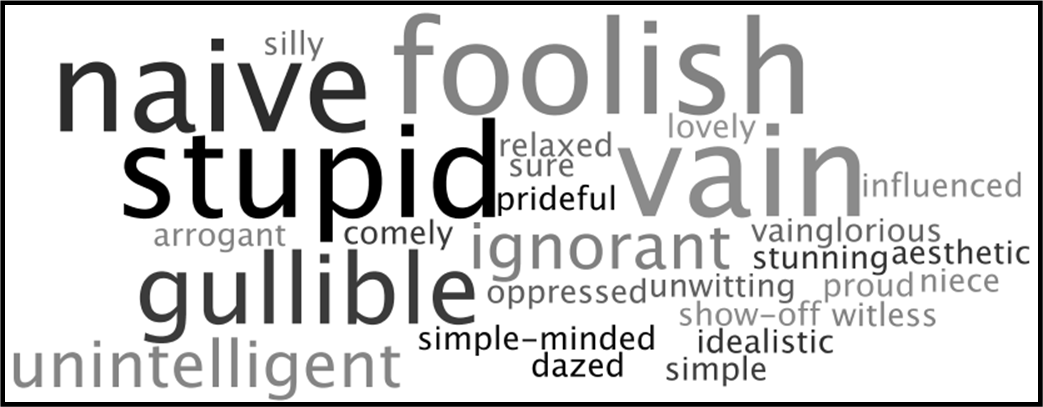}
                \caption{Crow neutral}
                \label{cloud-crow-neutral}
        \end{subfigure}
	\begin{subfigure}[b]{0.3\textwidth}
                \includegraphics[width=\textwidth]{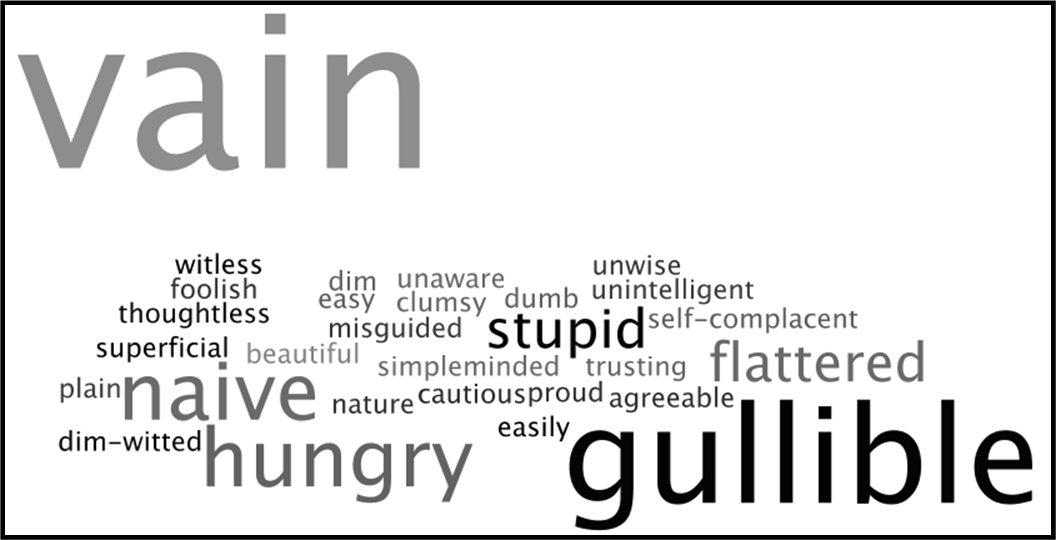}
                \caption{Crow Shy}
                \label{cloud-crow-shy}
        \end{subfigure}
        \begin{subfigure}[b]{0.3\textwidth}
                \includegraphics[width=\textwidth]{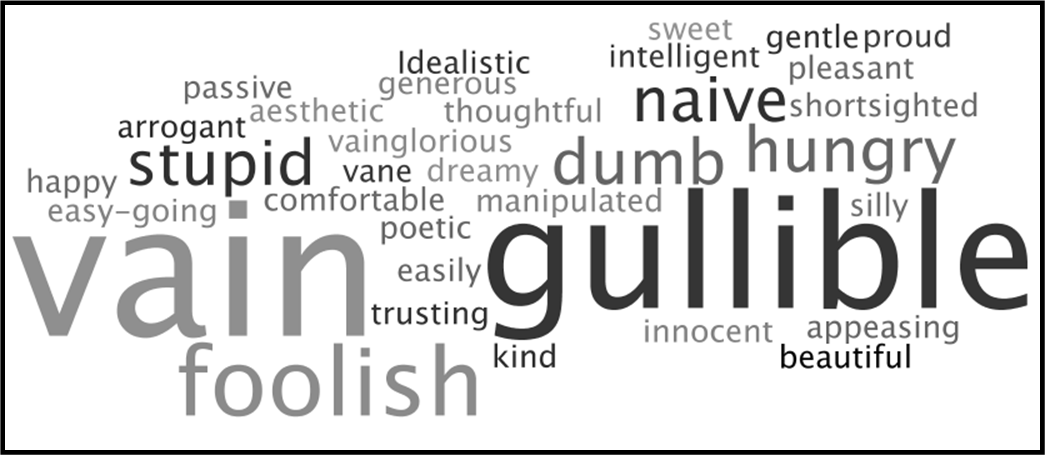}
                \caption{Crow Friendly}
                \label{cloud-crow-friendly}
        \end{subfigure}
        \caption{Word Cloud for the Crow}\label{cloud-crow}
\end{figure*}

\begin{figure*}[tbh]
        \centering
        \begin{subfigure}[b]{0.3\textwidth}
                \includegraphics[width=\textwidth]{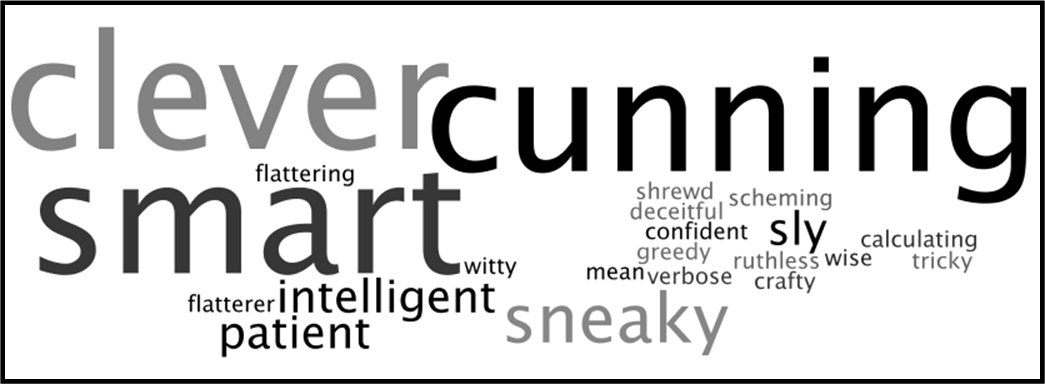}
                \caption{Fox neutral}
                \label{cloud-fox-neutral}
        \end{subfigure}
	\begin{subfigure}[b]{0.3\textwidth}
                \includegraphics[width=\textwidth]{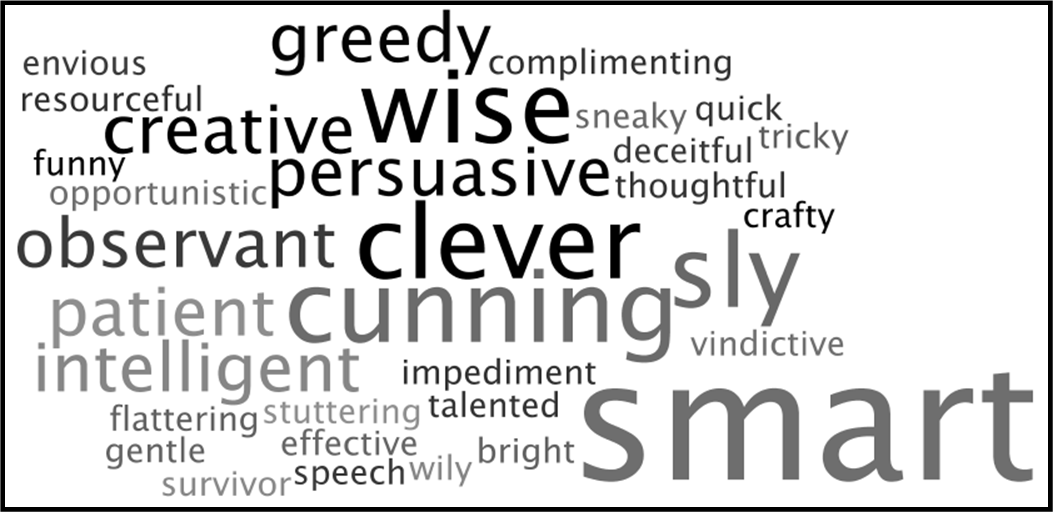}
                \caption{Fox Shy}
                \label{cloud-fox-shy}
        \end{subfigure}
        \begin{subfigure}[b]{0.3\textwidth}
                \includegraphics[width=\textwidth]{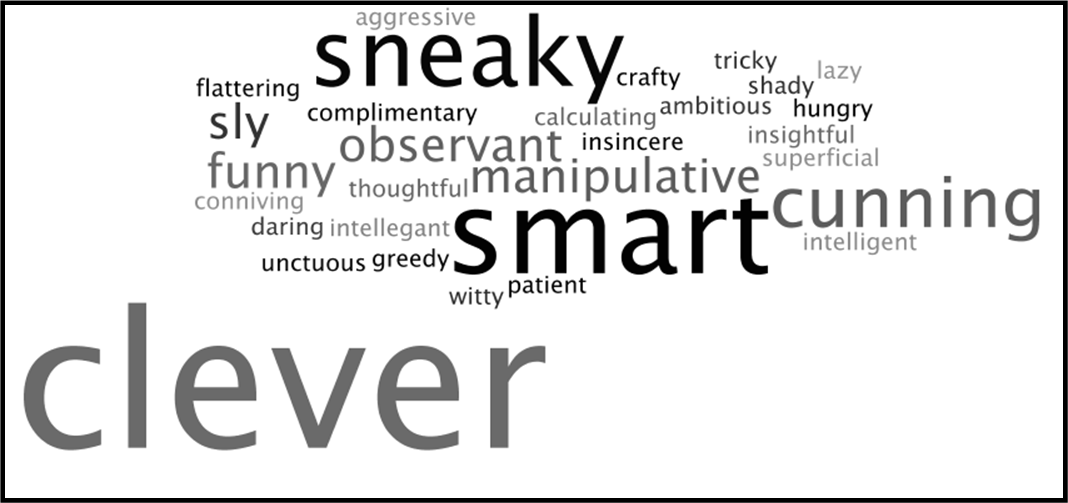}
                \caption{Fox Friendly}
                \label{cloud-fox-friendly}
        \end{subfigure}
        \caption{Word Cloud for the Fox}\label{cloud-fox}
\end{figure*}

After reading one of these tellings, we ask participants to provide adjectives in free-text describing the characters in the story. Fig.s~\ref{cloud-crow} and~\ref{cloud-fox} show word clouds for the adjectives for the crow and the fox respectively. The {\it shy} fox was not seen as very ``clever" or ``sneaky" whereas the {\it laid-back} and {\it neutral} fox were. However, the {\it shy} fox was described as ``wise" and the {\it laid-back} and {\it neutral} were not. There are also more positive words, although of low frequency, describing the {\it shy} fox. We observe that the {\it laid-back} and {\it neutral} crow are perceived more as ``na\"ive" than ``gullible", whereas {\it shy} crow was seen more as ``gullible" than ``na\"ive". {\it Neutral} crow was seen more as ``stupid" and ``foolish" than the other two models.

Table~\ref{crow-eval} shows the percentage of positive and negative descriptive words defined by the LIWC \cite{pennebaker2001linguistic}. We observe a difference between the use of positive words for {\it shy} crow and {\it laid-back} or {\it neutral}, with the {\it shy} crow being described with more positive words. We hypothesize that the stuttering and hesitations make the character seem more meek, helpless, and tricked rather than the {\it laid-back} model which is more boisterous and vain. However, there seems to be less variation between the fox polarity. Both the stuttering {\it shy} fox and the boisterous {\it laid-back} fox were seen equally as ``cunning" and ``smart".

This preliminary evaluation shows that there is a perceived difference in character voices. Furthermore, it is easy to change the character models for the {\sc est} to portray different characters. 

\begin{table}
\centering
\caption{\label{crow-eval} Polarity of Adjectives describing the Crow and Fox (\% of total words)}
\begin{small}
\begin{tabular}{|r|c|c||r|c|c|} \hline
Crow  & Pos & Neg & Fox & Pos & Neg \\ \hline
Neutral & 13 & 29 & Neutral & 38 & 4 \\
Shy & 28 & 24 & Shy & 39 & 8 \\
Laid-back & 10 & 22 & Laid-back & 34 & 8 \\ \hline
\end{tabular}
\end{small}
\end{table}

\section{Conclusion}

In this paper, we build on our previous work on the {\sc est} 
\cite{rishes2013generating}, and explain how it can be used to
allow linguistically na\"ive
authors to automatically generate dialogue variants of stock
utterances. We describe our extensions to the {\sc est} to handle direct
speech and interlocutors in dialogue.  We  experiment with
how these dialogue variants can be realized utilizing parameters for
characters in dynamic open worlds. \cite{walker2013using} generate
utterances using {\sc personage} and require authors to select and
edit automatically generated utterances for some scenes. A similar
revision method could be applied to the output of the {\sc est}.

As a potential future direction, we aim to explore the potential of
applying this approach to games with expansive open worlds with
non-player characters (NPCs) who come from different parts of the
world and have varied backgrounds, but currently all speak the same
dialogue in the same way. While above we discuss how our method could
be used to generate dialogue that varies according to character
personality, the {\sc est} could also be used to produce dialogue
variants corresponding to in-game regional dialects. {\sc personage}
models are not restricted to the Big Five personality traits, but
rather comprise values for 67 parameters, from which models for unique
regional dialects could easily be sculpted. Toward this,
\cite{walker2013using} created a story world called {\it Heart of
  Shadows} and populated it with characters with unique character
models. They began to create their own dialect for the realm with
custom hedges, but to date the full flexibility of {\sc personage} and
its 67 parameters has not been fully exploited. Other recent work has
made great strides toward richer modeling of social-group membership
for virtual characters \cite{harrell}. Our approach to 
automatically producing linguistic
variation according to such models would greatly enhance the impact
of this type of modeling. \\
\\
\noindent{\bf Acknowledgments} This research was supported by NSF Creative IT program grant \#IIS-1002921, and a grant from the Nuance Foundation.

\bibliographystyle{aaai}

\end{document}